\documentclass[letterpaper, 10 pt, conference]{ieeeconf}
\IEEEoverridecommandlockouts
\usepackage{cite}
\usepackage{amsmath,amssymb,amsfonts}
\usepackage{algorithmic}
\usepackage{graphicx}
\usepackage{textcomp}
\usepackage{xcolor}
\usepackage{csquotes}
\usepackage[utf8]{inputenc}
\usepackage[T1]{fontenc}
\usepackage{newtxtext, newtxmath}
\usepackage{hyperref}
\usepackage[capitalize]{cleveref}

\usepackage{float}
\usepackage{subcaption}
\usepackage{pifont}
\usepackage{longtable}
\usepackage{booktabs}
\usepackage{multirow}
\usepackage{threeparttable} 
\usepackage{lscape}
\def\BibTeX{{\rm B\kern-.05em{\sc i\kern-.025em b}\kern-.08em
    T\kern-.1667em\lower.7ex\hbox{E}\kern-.125emX}}
\begin{document}

\title{\LARGE \bf First Responders' Perceptions of Semantic Information for Situational Awareness in Robot-Assisted Emergency Response\\
}

\author{Tianshu Ruan$^{1}$, Zoe Betta$^{2}$, Georgios Tzoumas$^{3}$, Rustam Stolkin$^{1}$ and  Manolis Chiou$^{4}$
\thanks{$^{1}$Tianshu Ruan, and Rustam Stolkin are with Extreme Robotics Lab (ERL) and  National Center for Nuclear Robotics (NCNR), University of Birmingham, UK.}%
\thanks{$^{2}$Zoe Betta is with Department of Computer Science, Bioengineering, Robotics and Systems Engineering, University of Genova, Italy.}%
\thanks{$^{3}$Georgios Tzoumas is with School of Engineering Mathematics and Technology, University of Bristol, UK.}%
\thanks{$^{4}$Manolis Chiou is with School of Electronic Engineering and Computer Science, Queen Mary University of London, UK.}%
}

\maketitle

\vspace{-2ex}
\small
© 2025 IEEE. Personal use of this material is permitted. Permission from IEEE must be obtained for all other uses, in any current or future media, including reprinting/republishing this material for advertising or promotional purposes, creating new collective works, for resale or redistribution to servers or lists, or reuse of any copyrighted component of this work in other works.

\vspace{2ex}
\begin{abstract}

This study investigates First Responders' (FRs) attitudes toward the use of semantic information and Situational Awareness (SA) in robotic systems during emergency operations. A structured questionnaire was administered to 22 FRs across eight countries, capturing their demographic profiles, general attitudes toward robots, and experiences with semantics-enhanced SA. Results show that most FRs expressed positive attitudes toward robots, and rated the usefulness of semantic information for building SA at an average of 3.6 out of 5. Semantic information was also valued for its role in predicting unforeseen emergencies (mean 3.9). Participants reported requiring an average of 74.6\% accuracy to trust semantic outputs and 67.8\% for them to be considered useful, revealing a willingness to use imperfect but informative AI support tools.

To the best of our knowledge, this study offers novel insights by being one of the first to directly survey FRs on semantic-based SA in a cross-national context. It reveals the types of semantic information most valued in the field-such as object identity, spatial relationships, and risk context-and connects these preferences to the respondents' roles, experience, and education levels. The findings also expose a critical gap between lab-based robotics capabilities and the realities of field deployment, highlighting the need for more meaningful collaboration between FRs and robotics researchers. These insights contribute to the development of more user-aligned and situationally aware robotic systems for emergency response.
\end{abstract}


\section{Introduction}
Extreme environments such as disaster zones, radiation-exposed areas, or structurally unstable buildings pose serious risks to human life. In these situations, First Responders (FRs) are often required to operate under high stress, with limited visibility, communication challenges, and physical danger. \par
At the same time, robotic technologies — such as drones, ground vehicles, and other autonomous systems — are becoming increasingly accessible and capable, due to rapid advances in perception, planning, and control. Commercial applications are expanding, yet these developments have not fully translated into effective tools for FRs operating in real-world extreme conditions. Despite growing interest, a significant gap persists between laboratory robotics and field deployments. Challenges such as rugged terrain, environmental hazards (e.g., heat or radiation), and unreliable communications continue to limit the usefulness of robots in the field \cite{huang2023went}. \par
As robotic technology matures, it becomes increasingly important to ensure its development is informed by the needs and experiences of end-users in the field, especially FRs. Their input is critical in designing systems that can truly support them in the field. In many cases the input of firefighters has been integrated at early stages of the development of these technologies, but in some cases this is not always performed \cite{10558013, mcconville2024adoption}. \par
To help bridge this gap, we conducted a structured survey aimed at gathering feedback from FRs and robot operators across different mission types and backgrounds. Our methodology builds on previous work \cite{ruan2022taxonomy}, where we developed a taxonomy of semantics relevant to robotics, and in this study, we presented these concepts to FRs to gather their opinions. We asked them how semantic information, like social media-based Situational Awareness (SA) \cite{snyder2019situational}, might support robotic deployments in real-world operations. \par
The aim of this paper is twofold. First, we explore the perspectives of FRs on the challenges and demands they face when using robots in the field. Second, we investigate how our proposed semantic framework aligns with their experiences and needs. By analyzing their responses, we hope to better understand the practical role of semantic information in field robotics and identify how future research can be directed to support the real-world demands of FRs.

\section{Related work}
This work aims at investigating SA, in the context of Search and Rescue (SAR). SA has been defined by Endlsey \cite{endlsey1995} as three levels - ``\textbf{i)} the perception of the elements in the environment within a volume of time and space, \textbf{ii)} the comprehension of their meaning, and \textbf{iii)} the projection of their status in the near future''. 

SA is a well-defined concept that encounters significant challenges whenever researchers try to measure the level acquired during operations \cite{Cooper2013-bq}. The different techniques can be direct or indirect to measure different aspects of SA. This highlights a challenge in objectively determining whether a product guarantees an SA level that is adequate for the task that is to be completed. When talking about dynamic SA in the scope of human-robot teams, \cite{senaratne2025framework} systematically discussed the dynamic nature of team SA and the factors that affect SA.

In disaster management, the problem of acquiring SA has been analyzed by several researchers. Some works propose solutions to improve SA in different applications. For example, the work from Allison et al. \cite{allison2024} proposes a framework to control different robots in a hazardous environment with the objective of collecting information and improving the decision-making process. Similarly, in the work proposed by Yang et al. \cite{yang2014}, the goal was to develop a robot able to aid in the SAR task by increasing the information collected, and consequently the SA acquired during the mission. In both of these works, the focus is on the engineering side of the solution and not on how the solution improves the SA of the end users. The effect on the SA is not measured, and the opinion of the end users has not been formally collected.

Other works, such as \cite{betta2024perceptions} and \cite{murphy2025rural}, focus more on the human aspect of the task. The work detailed in \cite{betta2024perceptions} focuses on the requirements of FRs for robotic deployment in a SAR scenario. On the other hand, the work from Murphy et al. \cite{murphy2025rural} focuses on the collection of the challenges in rural areas for what concerns disaster management operations. Both of these works propose a series of interviews with experts in the field to gather information from FR experts. 
In a similar way, with our work, we propose to investigate how FRs acquire SA and if they would be open to using robots with semantic information to improve the decision-making process.

\section{Methodology}
In this survey, insights and demands were collected via online questionnaires and paper questionnaire sheets. To reach field experts in different countries, the questionnaires were translated from English into Italian, Greek, and Chinese by members of our multi-national research group, by using the back-translation technique. 
This helps to minimize any misunderstandings caused by second-language users, particularly given the abstract concepts in the questionnaire. The questionnaire consists of three sections: \textbf{Background and basic demographic information}, \textbf{General attitudes towards robots}, and \textbf{Situational awareness and use of semantic information}. 

In the questionnaires, we investigate the FRs' backgrounds from two key dimensions: personal information and experiences with robots. 
The demographic characteristics include the following features: age, gender, country, education level, roles in the team/field, experience in the field, experience with robots, and others. Next, we propose a validated questionnaire, General Attitudes Towards Robots (GAToRS) \cite{koverola2022general}, as they may affect how FRs perceive domain-specific robotics. The Italian version of the GAToRS has been translated following the validated translation provided by Carradore et al. \cite{Carradore2024}. 
In the questionnaire section \textbf{Situational awareness and use of semantic information}, we survey FRs about their usage, experience, expectations, and overall opinions on SA and semantics.  

By leveraging our professional networks and contacts, we successfully invited 22 FRs from eight countries to participate in the survey. All FRs are safety and emergency professionals with relevant experience, covering multiple countries and emergency roles, e.g. firefighters, emergency medical personnel, technical operators, on-site commanders, operational scientists, vehicle drivers, and others.

\section{Results}
\subsection{FR background and demographics}
\begin{figure}[htbp]
    \centering

    \begin{subfigure}[b]{0.44\textwidth}
        \centering
        \includegraphics[width=\textwidth]{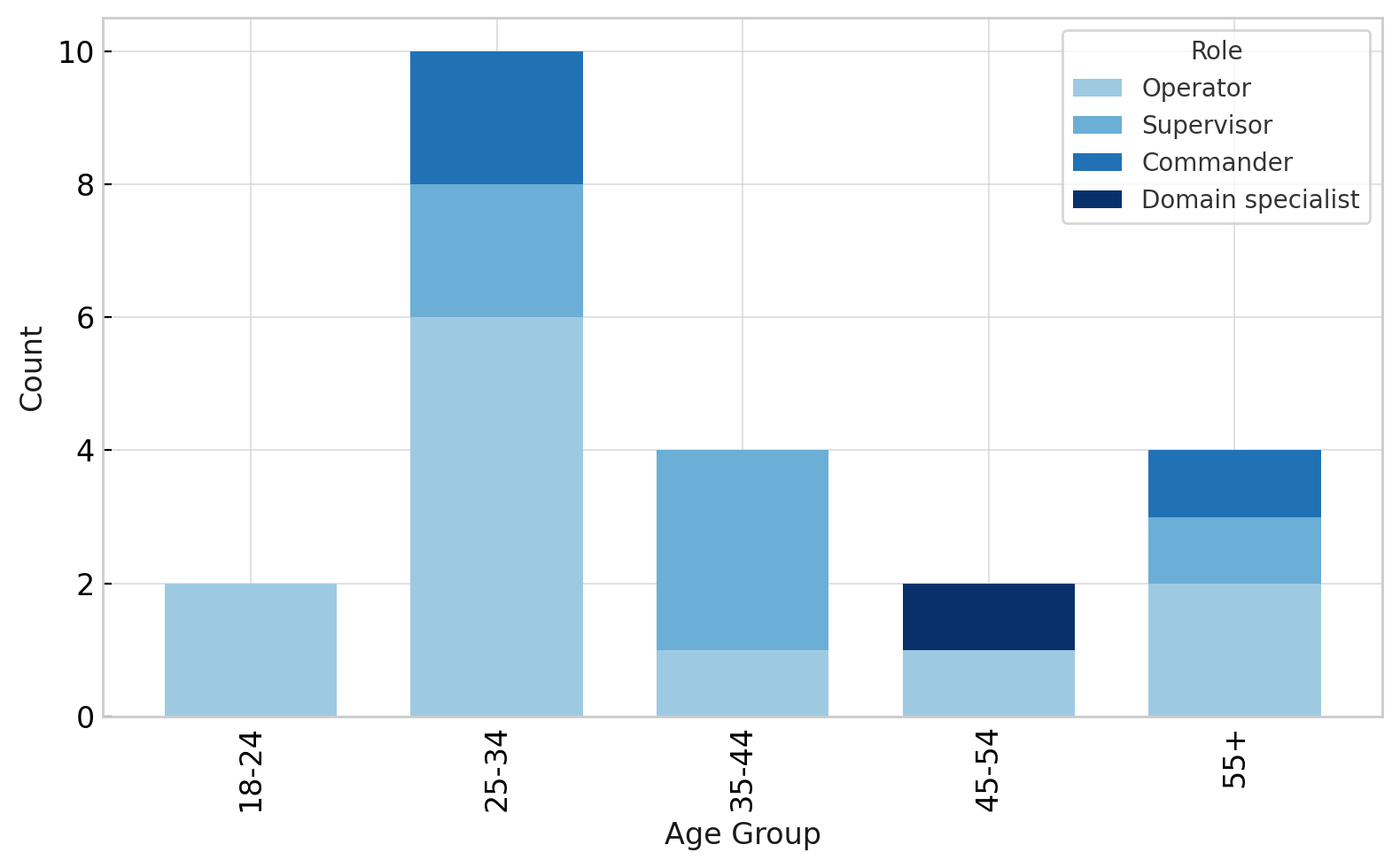}
        \caption{Role distribution by age.}
    \end{subfigure}
    \hfill
    \begin{subfigure}[b]{0.44\textwidth}
        \centering
        \includegraphics[width=\textwidth]{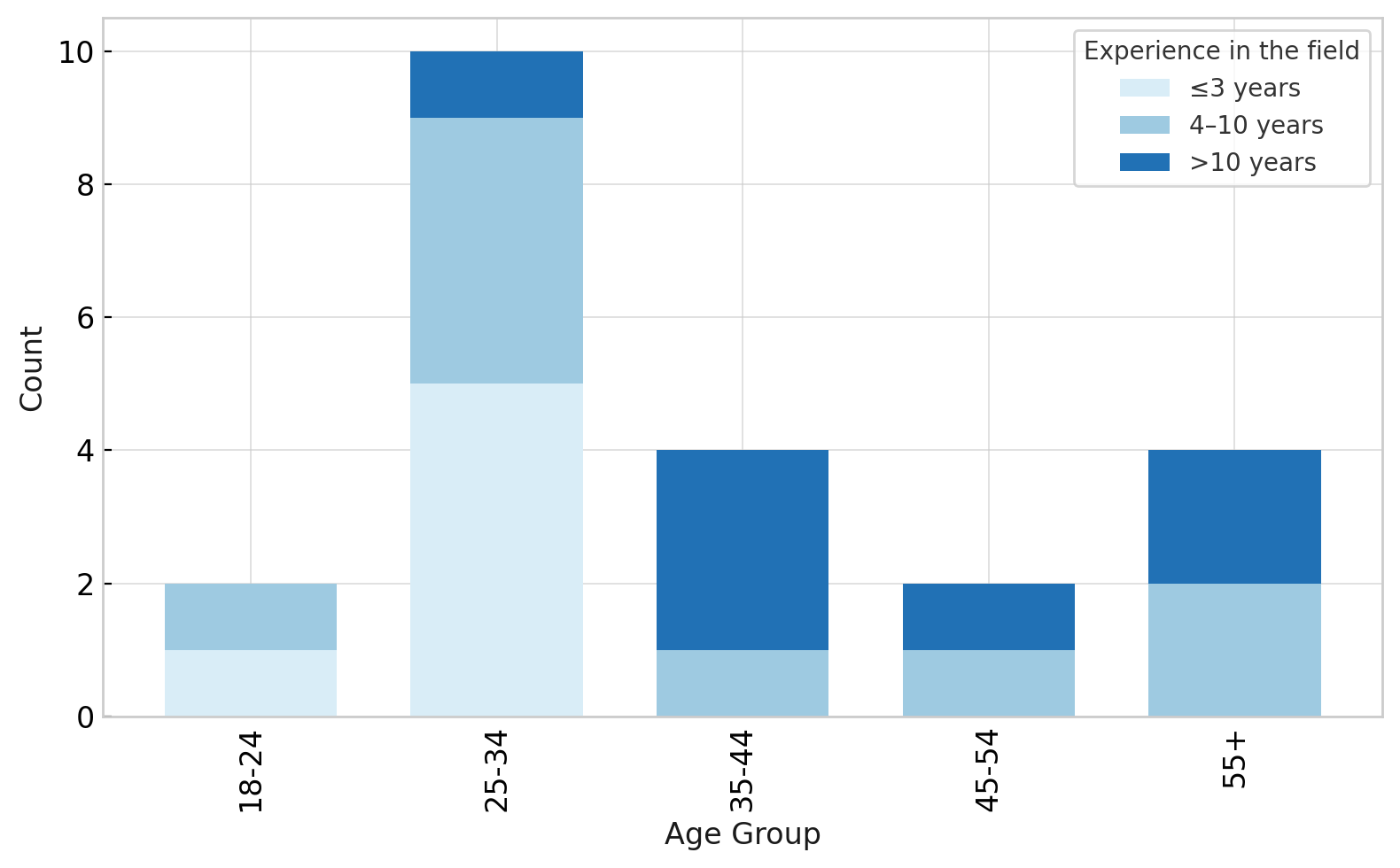}
        \caption{Field experience distribution by age.}
    \end{subfigure}
    
    \begin{subfigure}[b]{0.44\textwidth}
        \centering
        \includegraphics[width=\textwidth]{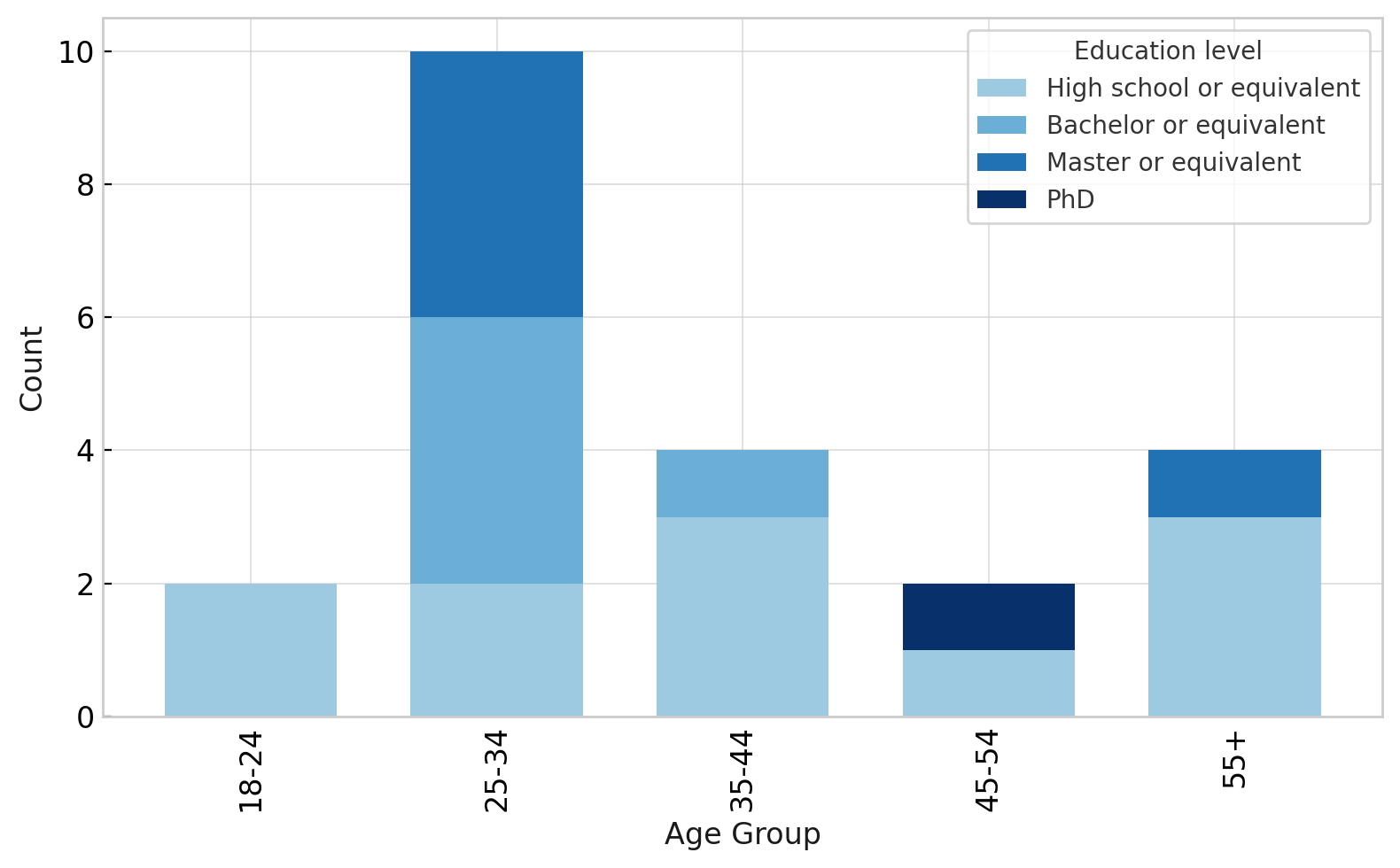}
        \caption{Education distribution by age.}
    \end{subfigure}
    \hfill
    \begin{subfigure}[b]{0.44\textwidth}
        \centering
        \includegraphics[width=\textwidth]{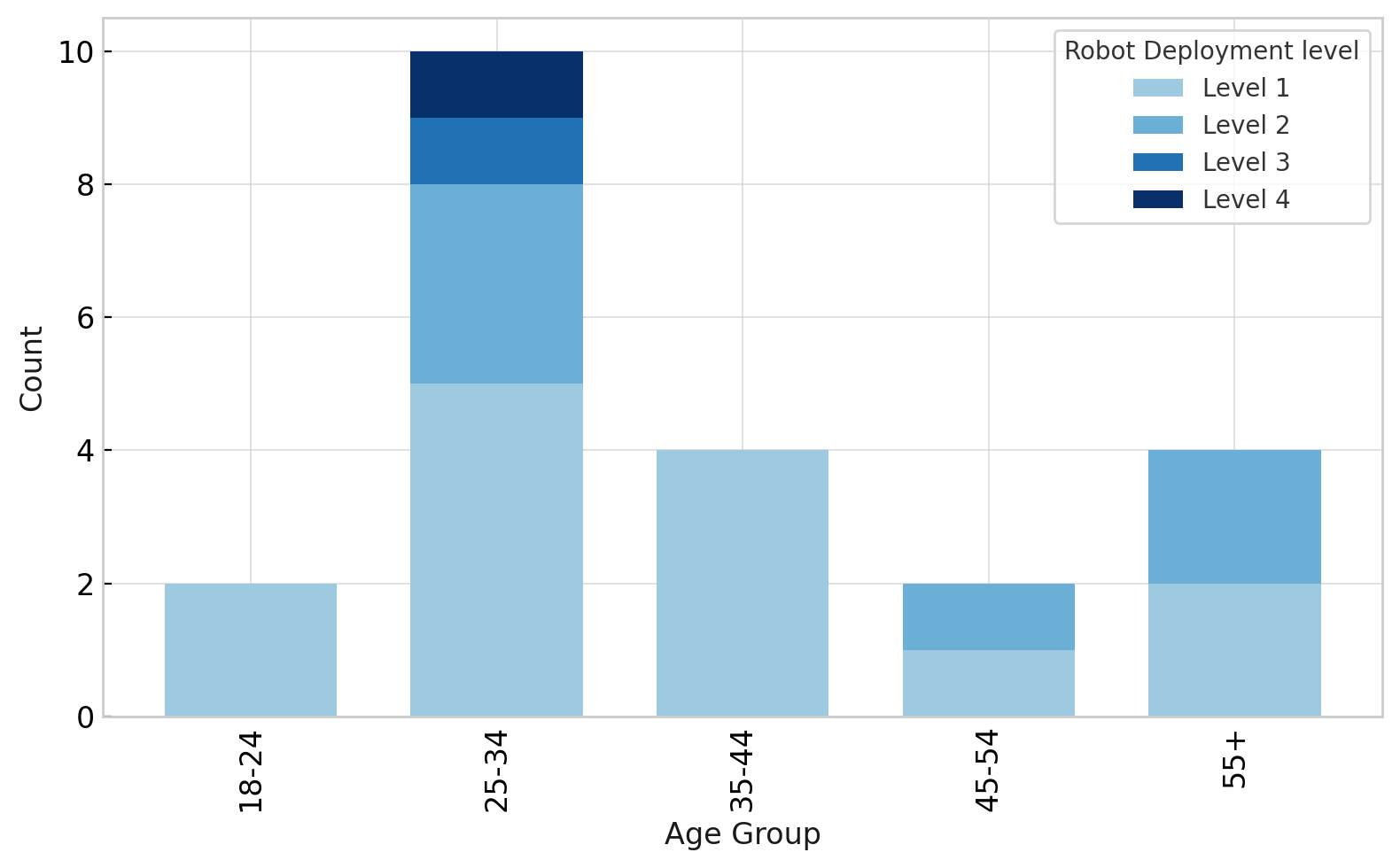}
        \caption{Robot deployment level distribution by age.}
    \end{subfigure}

    \caption{Distributions of role, field experience, education level, and robot deployment level by age group. Robot deployment level refers to the frequency of using robots in the mission. Higher level means higher frequency.}
    \label{fig:background-distributions}
\end{figure}

The recruited FRs are from: China, Japan, Germany, Iraq, Italy, Greece, the UK, and the USA. We understand there might be bias caused by the number of FRs from different regions. However, it is difficult to reach the FRs all over the world, let alone equal numbers. Fortunately, we noticed they shared a similar standard of process and training in the field. We believe the bias caused by the unequal number would not affect the findings much. From the demographics, we notice females are much less represented than males. This is likely due to the heavy physical demands involved in many FR roles, which still tend to be male-dominated in most countries. Only three female FRs participated in our study. Similarly, for the age distribution, FRs below 35 years old make up the half of participants in the survey. They mostly work on the front line of missions, while older FRs are more senior in the field, serving as commanders or scientists who give expert instructions and guidance. For the same reason, FRs do not typically require a very high level of academic education. More than 70\% of FRs in the survey achieved a degree below master's or equivalent. Even some senior commanders do not have a higher level of education. However, they do require special training, skills, experience, and knowledge to tackle different types of emergencies. When talking about their expertise and experience, FRs show a balanced distribution. The junior (<3 years experience), intermediate (4 - 10 years experience), and senior (>10 years experience) FRs take 27\%, 42\%, and 31\% respectively. In addition, we inquired about their robotics experience and the current robot deployment status from their perspective in the field. Important distributions can be seen in \cref{fig:background-distributions}. 

From the distributions in \cref{fig:background-distributions}, we notice that there are relationships between certain background factors, such as field experience and age. Even though there are some unusual cases, e.g. senior FR personnel with only a high-school education level, the distribution reveals some patterns. Older FRs are more experienced, tend to be better educated, and typically take leadership roles when compared with younger FRs. However, the opinions of both older and younger FRs on current robot deployments in the field converge. Most FRs feel that robot deployments are still not very advanced, and few of the deployed robots can operate fully autonomously. 

\subsection{General attitudes towards robots}
\label{attitudes towards robots}
As robots continue to grow in prevalence in society, people have developed different perceptions and attitudes towards them. Such background attitudes might potentially become a confounding factor in our survey, when we ask about more specific opinions on the deployment of robots in FR and extreme environment applications. We therefore first investigate and analyze participants' general attitudes towards robots as a reference, and then explore their attitudes towards robots in their own specialized application areas.

The general attitude section is designed based on \cite{koverola2022general}. The validated questionnaire is the General Attitudes Towards Robots Scale (GAToRS). The scale differentiates \textbf{i)} comfort and enjoyment around robots, \textbf{ii)} unease and anxiety around robots, \textbf{iii)} rational hopes about robots in general (at societal level), and \textbf{iv)} rational worries about robots in general (at societal level) \cite{koverola2022general}. Hence, the questionnaire is divided into four dimensions: 

\begin{itemize}
\item\textbf{~Personal level positive attitude (P+)} evaluates personal positive feelings and emotions toward robots, e.g. relaxed, trust, and friendly interaction. 

\item\textbf{~Personal level negative attitude (P-)} evaluates personal negative feelings and emotions toward robots, e.g. anxiety, upset, and fear.

\item\textbf{~Societal level positive attitude (S+)} evaluates expectations for the positive role of robots at the social level, e.g. assisting humans and removing humans from hazardous environments.

\item\textbf{~Societal level negative attitude (S-)} evaluates social concerns about the potential impact of robots, e.g. unemployment, regulations, and changes in social structure. 
\end{itemize}

Different statements are given to participants from the above dimensions. Participants need to assign a score that indicates to what extent they agree with the statements, e.g. ``I can trust a robot.'', ``Widespread use of robots is going to take away jobs from people.'' The scale is from one to seven, with one denoting ``totally disagree'' and seven denoting ``totally agree''.

Apart from these four dimensions, additional four statements are used to obtain a clearer picture of participants' general views. The additional statements are used to help validate whether the analysis drawn from the four dimensions is conflicting.

\begin{figure}[htbp]
  \centering

  \begin{subfigure}[b]{0.235\textwidth}
    \includegraphics[width=\textwidth]{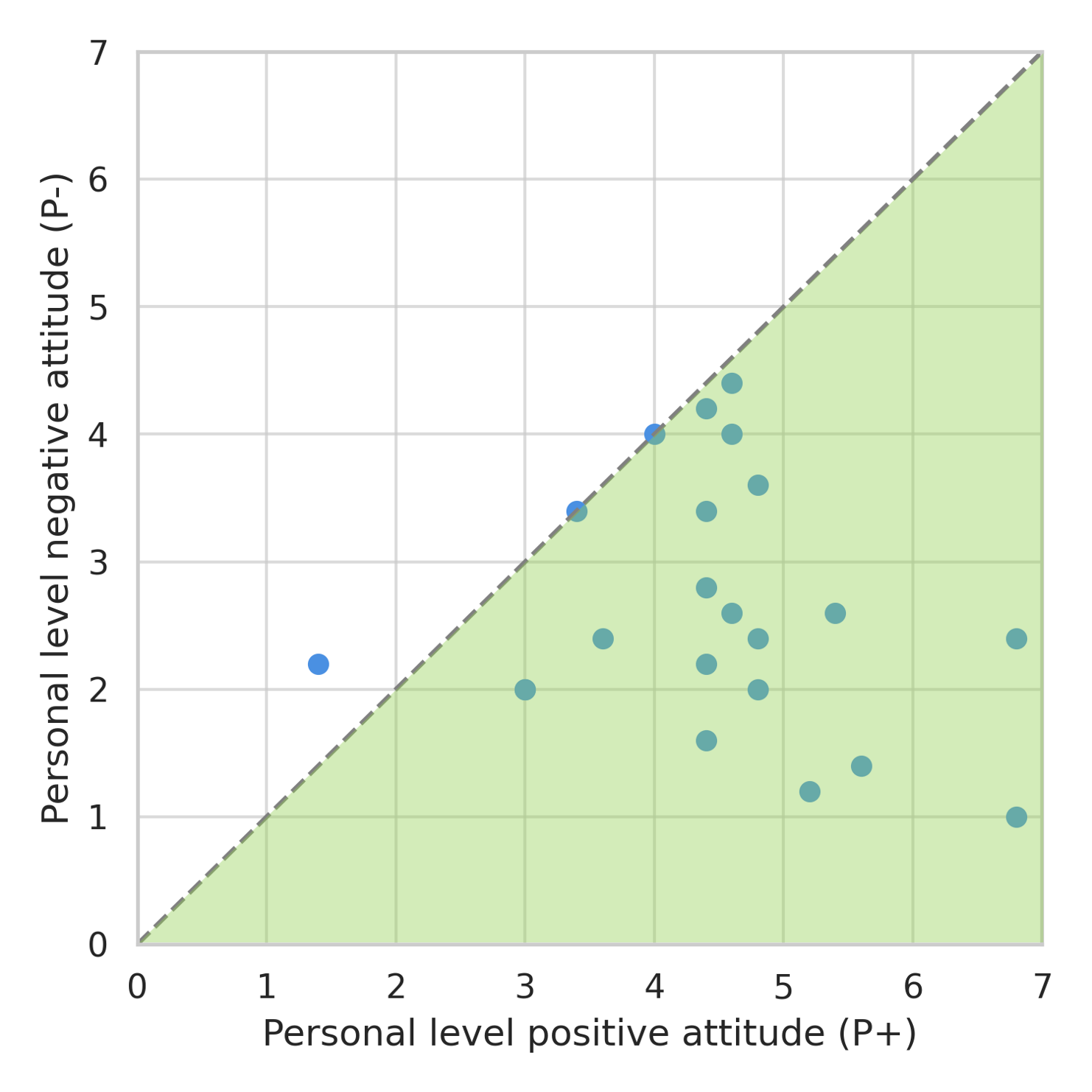}
    \caption{Personal Level Attitude}
    \label{fig:personal}
  \end{subfigure}
  \begin{subfigure}[b]{0.235\textwidth}
    \includegraphics[width=\textwidth]{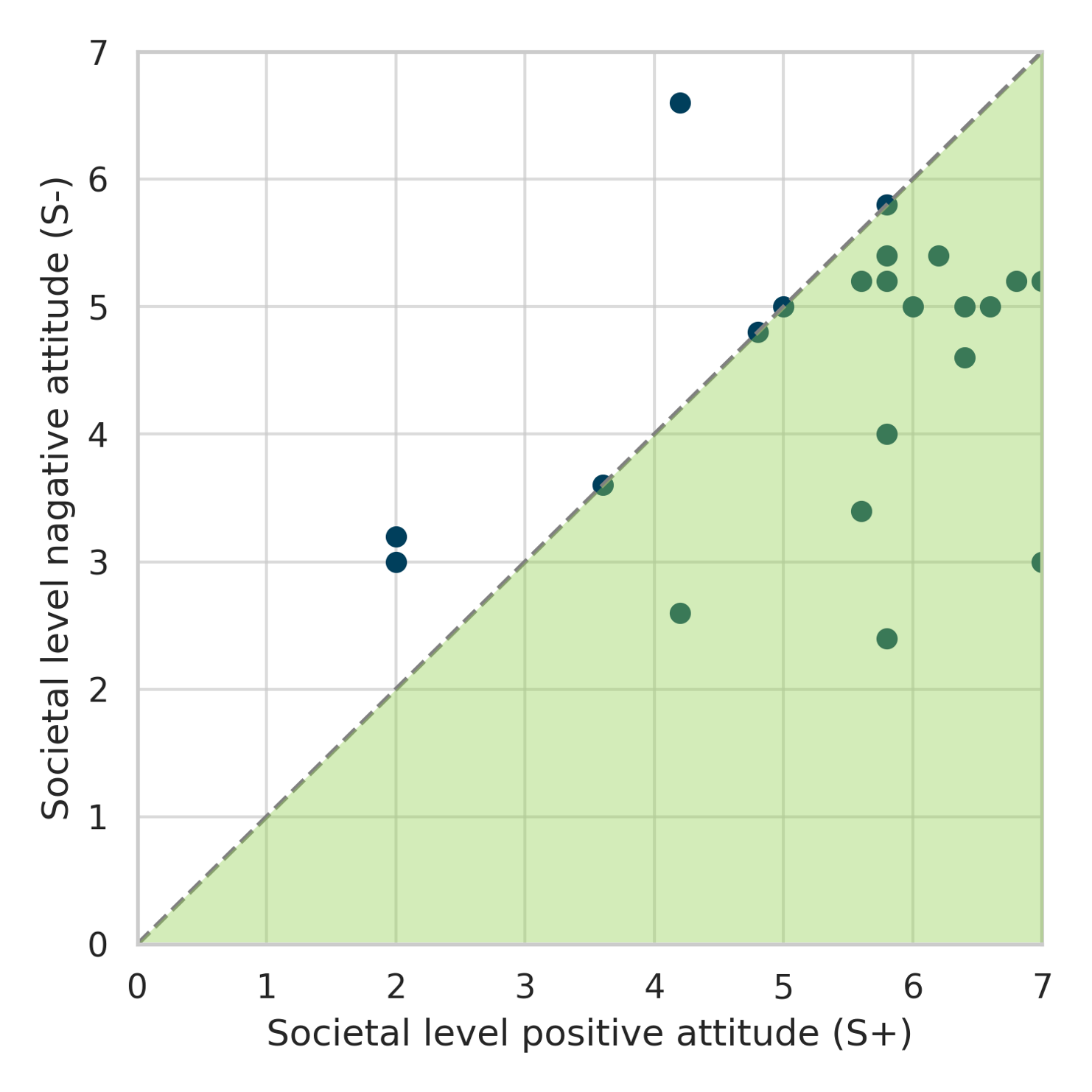}
    \caption{Societal Level Attitude}
    \label{fig:societal}
  \end{subfigure}

  \caption{Distribution of general attitudes towards robots on personal and societal levels. Most points are under the diagonal line (in the green zone). This indicates that most FRs have positive attitudes towards robots in both personal and societal contexts.}
  \label{fig:attitudes}
\end{figure}

The distributions of FR's attitudes are presented in \cref{fig:attitudes}. The y-axis refers to the negative view, and the x-axis refers to the positive view. In \cref{fig:personal}, most points are close to or below the diagonal line (in the green zone), where positive views are more than negative views. In other words, most of the FRs demonstrated more positive personal views than negative personal views. Only one FR who flies the drone in the field shows more negative personal attitudes than positive attitudes. Apart from them, most FRs indicate a certain degree of trust, relaxation, and welcome towards robots. At the same time, they have some concerns and reservations. However, these are fewer than their positive views. In \cref{fig:societal}, the participants show greater concern about robots at a societal level. Comparing with the \cref{fig:personal}, we notice that most points cluster in the upper right of the green area. This suggests that most participants are confident about the positive influence of robots on society. However, they are more concerned about the negative impacts of robots on society, than about the negative impact on individuals. Moreover, there are three FRs above the diagonal line, i.e. three participants (13\%) indicated stronger concerns about the impact of robots on society in general. In the end, the additional four statements about participants’ general views towards robots converge with the above four dimensions. In summary, most FRs are willing to trust and use robots without having very strong concerns about societal impact. It is apparent that the surveyed FRs are mostly pro-robots, which suggests that they might be willing to err on the side of giving the benefit of the doubt and showing more tolerance and patience to the introduction of robots in their missions. It is natural that they have a positive view of SA and the use of semantic information.

\subsection{Views about SA and use of semantic information}
In this section, we would like to know more about FR's views on SA and semantic information, based on their understanding and experience. We give some examples and briefly explain the definition of SA and semantics in this context. 

In this section, the numerical questions with a marking scale and open-ended questions are provided to obtain more specific ideas on different occasions, e.g. different task backgrounds. Specifically, the numerical  questions are listed below:

\begin{itemize}
\item How much do you have an estimation of the Situational Awareness before the mission? (Q43)

\item How often do you experience loss of concentration, fatigue, and cognitive overload? (Q47)

\item To what extent do you think having semantic information is useful in building Situational Awareness? (Q49)

\item To what extent is semantic understanding and reasoning used during missions? (Q50)

\item Can semantic information and semantic scene understanding help predict or mitigate an emergency that was not predicted or foreseen before the mission? (Q53)

\item Imagine you have some AI agent assisting in your mission (e.g. in a robot, in a computer, others). How much do you agree with the statement "It is helpful if the AI agent can process an understanding of the situation"? (Q54)
\end{itemize}

These questions are answered on a Likert scale \cite{joshi2015likert} from one to five, where one refers to a low level (e.g. totally disagree, low frequency etc.), and five refers to a high level (totally agree, high frequency etc.). 

\begin{figure}[htp]
    \centering
    \includegraphics[scale =0.43]{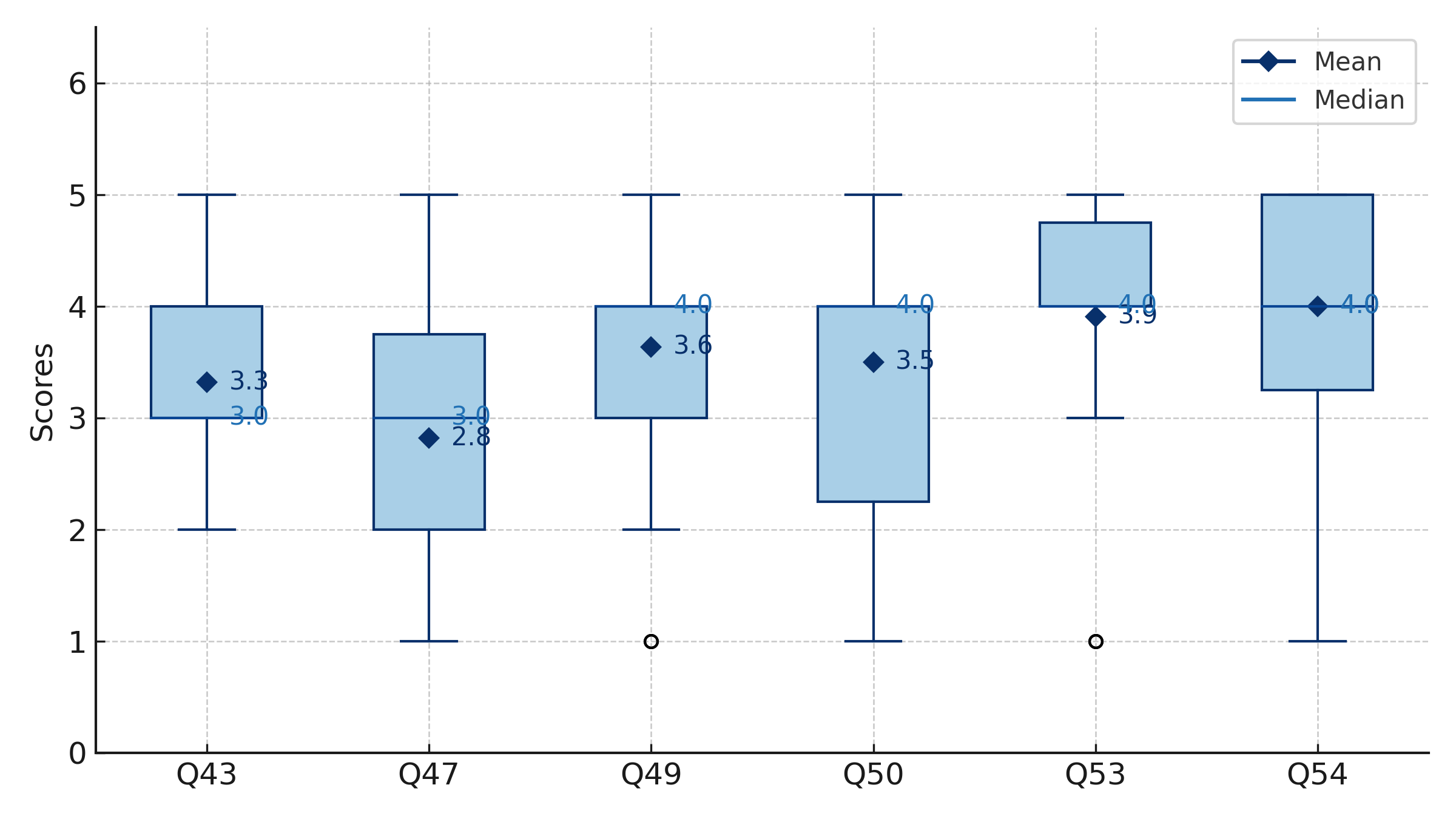} 
    \caption{Score distributions of numerical questions, including: SA level before mission (Q43); Over workload experience (Q47); Semantics help SA? (Q49); Semantics used level currently (Q50); Semantics help prediction? (Q53); If AI understands the situation, helpful? (Q54)}
    \label{fig:SA_semantics_mark}
\end{figure}

The results of the numerical questions are shown in \cref{fig:SA_semantics_mark}.

The open-ended questions covered a wide range of topics. They have two types of formats: fill-in questions and multiple choice (additional choices not listed are supported by adding after the last choice option). Below, we summarise the answers provided by participants. Where digits are given in brackets, these refer to the number of FRs who aligned with certain choices and options in their answers. Moreover, the answers to fill-in questions are presented with no preference, i.e. we do not rank the importance of the information from participants.

\begin{itemize}
\item\textbf{Information known before mission (Q42):} Type of event (location, hazardous material), extent, people/animal (distribution) involved; type of intervention required; specific task goals; weather forecasts; information from emergency call center that collected from people onsite; availability of resources (nearby medical center, airlifting services).

\item\textbf{Methods to obtain SA (Q44):} Camera/images (16); sensor readings (14);  communications with other people from the site, e.g. civilians, other team members (16); in-person exploration (16); comparison of current surveying data with previous status (e.g. Google maps) (1).

\item\textbf{Challenges to obtain SA (Q45):} Unstable and low speed communication with the sensors; communication delay among team members; malfunction of perception devices; rapid changes of environments (seemed a safe area but rapidly changed to dangerous); data integration; incomplete data; complex terrains and debris; low visibility; darkness; complex air traffic control and conflicts for drone deployment (e.g. drones interfering with helicopters); lack of sound; lack of 360 vision; false or inaccurate information.

\begin{figure*}[htp]
    \centering
    \includegraphics[scale =0.69]{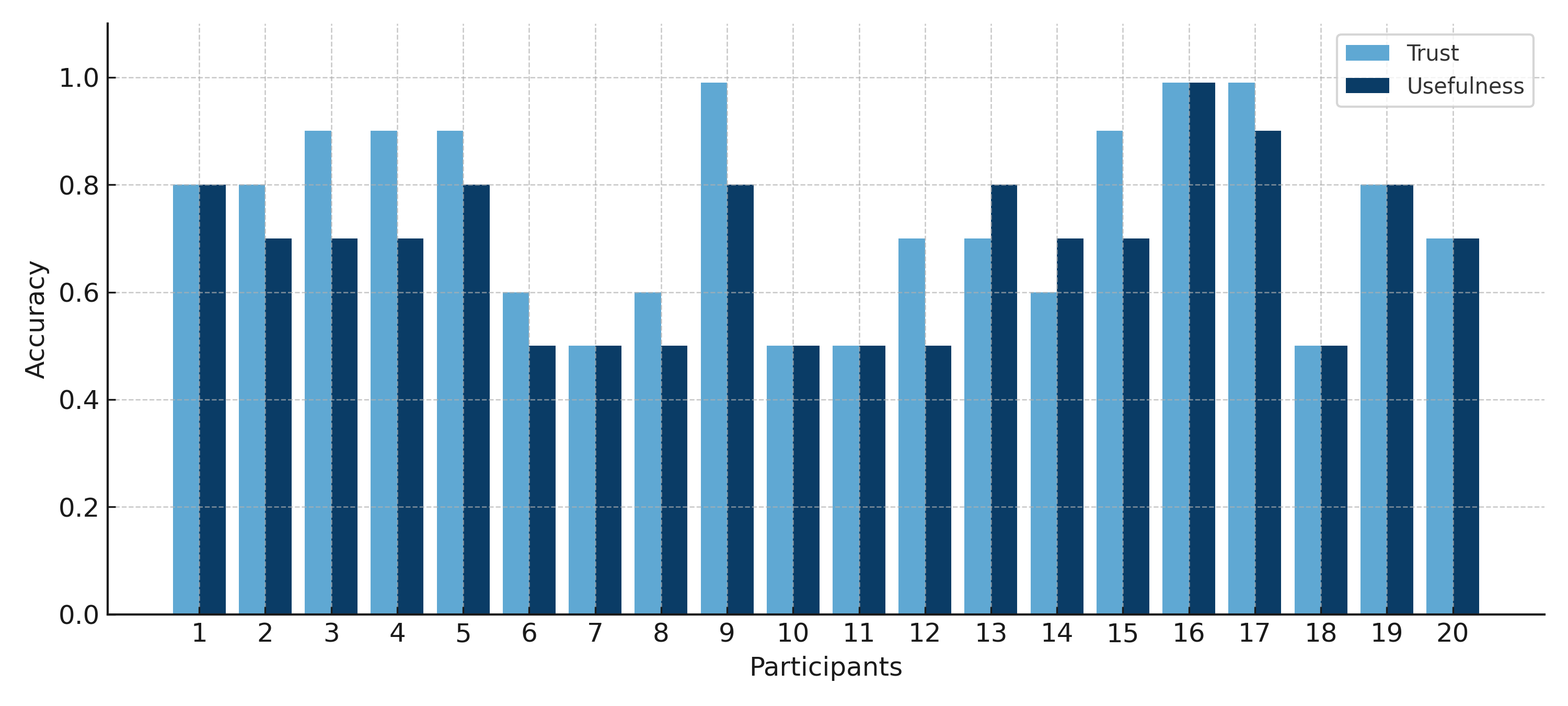}
    \caption{Accuracy threshold values for achieving trust and usefulness: light blue bars refer to the accuracy threshold for achieving trust, and dark blue bars refer to the accuracy threshold for achieving usefulness. Light blue bars are equal to or higher than dark blue bars for most FRs.}
    \label{fig:accuracy change}
\end{figure*}

\item\textbf{Top 3 important semantics (Q46):} Type of event (location, hazardous materials), extent, people/animal (distribution) involved; entities in the area (vehicles, victims, obstacles), impacts caused by interactions among entities (an ambulance blocked by an abandoned vehicle, risks from the hazardous materials); entities positioning in space and time (estimate path and schedule of an active ambulance); availability of resources (nearby medical center, airlifting services); object and hazard identification; task-relevant location information; distance to the object; available path.

\item\textbf{Reason of fatigue, and cognitive overload (Q48):} 24 hours work shift rotations; chaotic situations; difficulties in collaboration and communication within the team; simultaneous presence of heterogeneous and fragmented information; time pressure to make decisions in the circumstance of having incomplete or uncertain information; high responsibility and fear of error; complex environments (high temperature, potential hazards); infrastructure and technological shortage; physical fatigue after intensive non-stop work; heat stress due to wearing protective gear like Tyvek suits; victims; situation (injury, death); overload information; communication with many stakeholders; lose control of the robot; multi-tasking; conflicts between field commander's demand and safety regulation.

\item\textbf{Current method to deliver semantics (Q51):} Vision (20); sound (15); haptics (4); vibration (4); 

\item\textbf{Ideal method to deliver semantics (Q52):} Vision (21); sound (19); haptics (7); vibration (11); 
\end{itemize}

Finally, two questions about semantics accuracy were surveyed: the accuracy threshold for having trust in semantics; and the accuracy threshold for usefulness of semantics (See \cref{fig:accuracy change}). Only 20 participants gave a specific accuracy. The other two participants provided individual insights about this comparison. One participant thinks any accuracy is fine, as long as FRs are informed about the reliability of the accuracy. Another participant feels that the accuracy requirements vary depending on different occasions. But this participant feels that such sensors should show better performance than human perception and inference. 

\section{Analysis and discussion}
\begin{figure*}[htbp]
    \centering
    \includegraphics[width=0.89\textwidth]{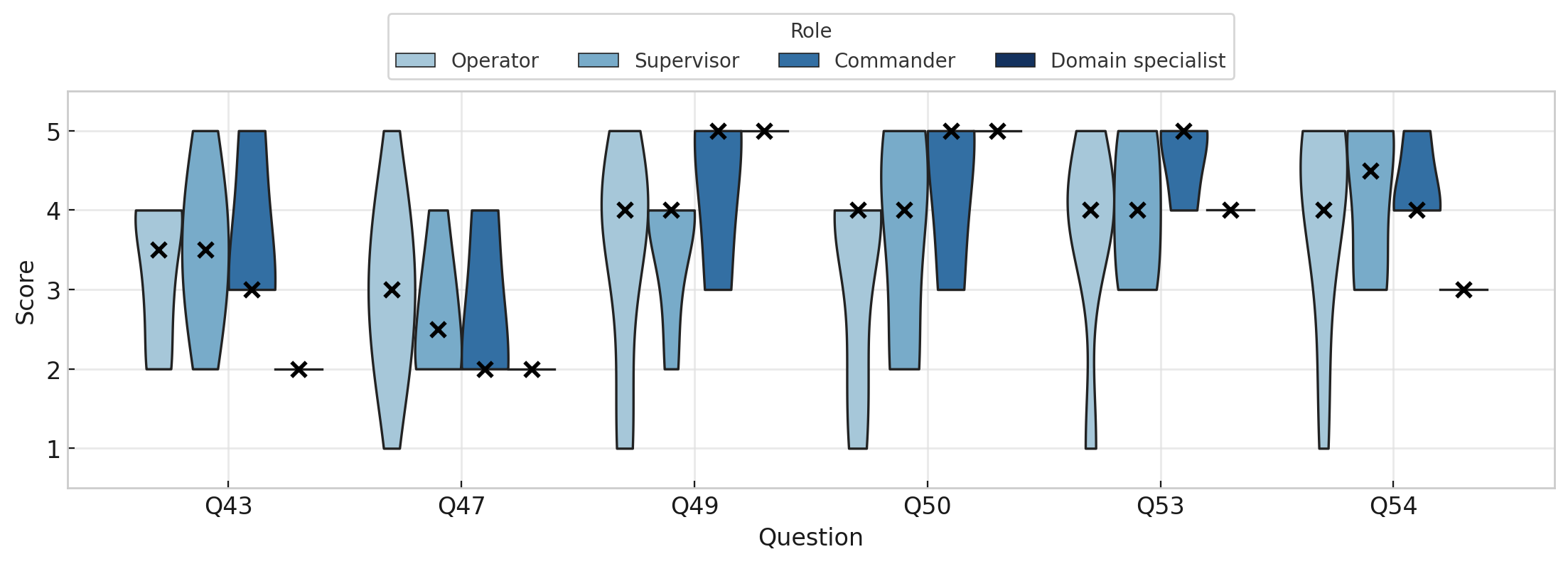}
    \caption{Score distribution by role: crosses refer to the median in the group. In each question, the roles displayed from left to right are operator, supervisor, commander, and domain specialist.}
    \label{fig: Score Distribution by Role}
    \vspace{0.5cm}  

    \includegraphics[width=0.89\textwidth]{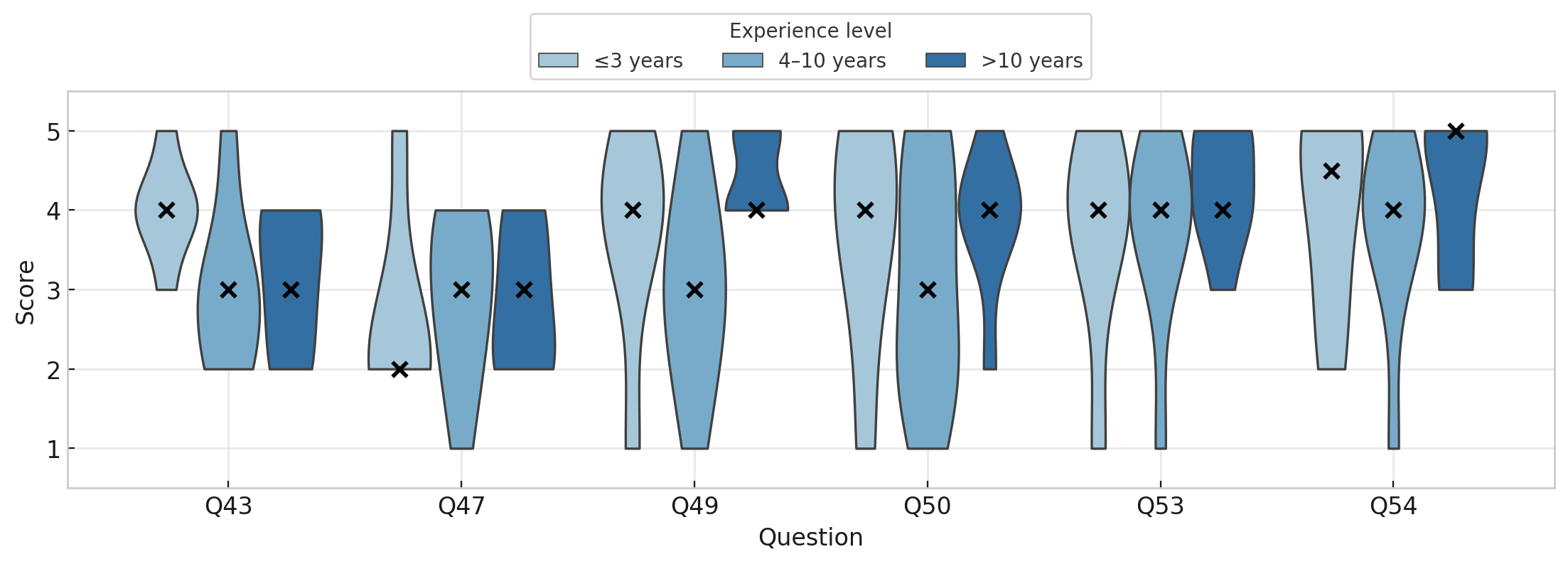}
    \caption{Score distribution by experience level: white crosses refer to the median in the groups and thick black lines represent the interquartile range. In each question, years of experience displayed from left to right are: less than 3 years, 4 to 10 years, +10 years.}
    \label{fig: Score Distribution by Experience Level}
    \vspace{0.5cm}

    \includegraphics[width=0.89\textwidth]{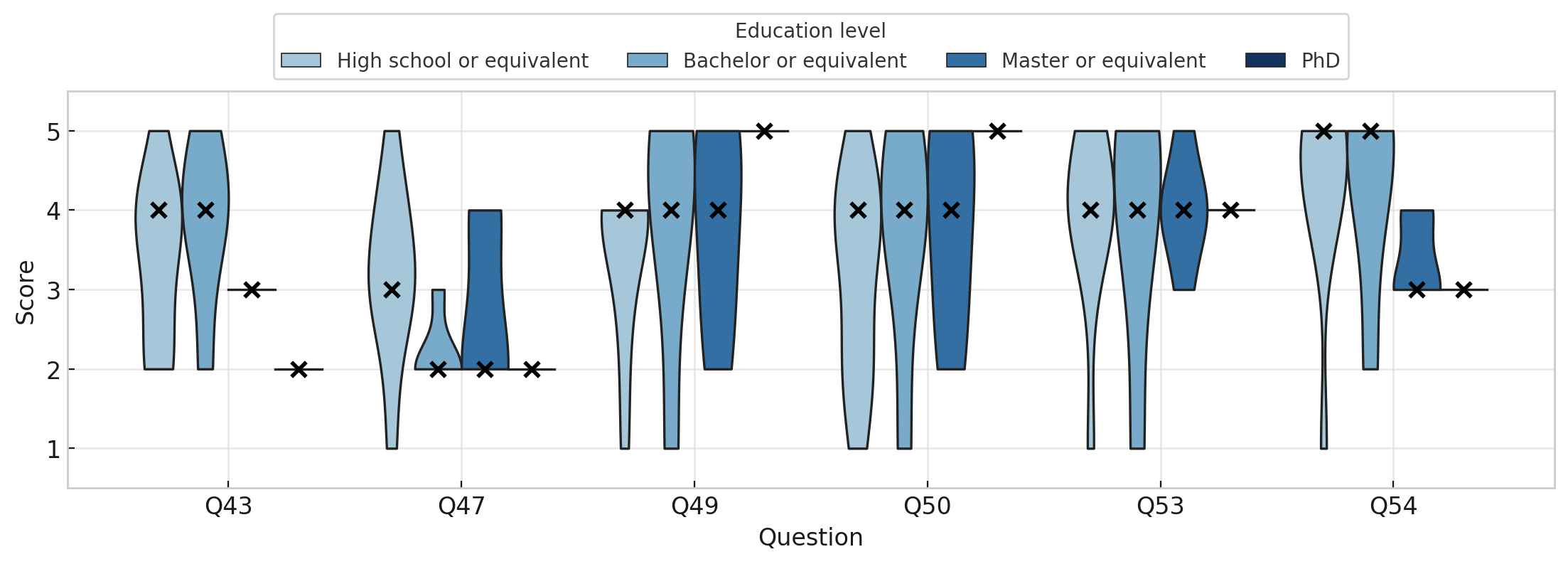}
    \caption{Score distribution by education level. In each question, the education level displayed from left to right is high school, bachelor, master, and PhD.}
    \label{fig: Score Distribution by Education Level}
\end{figure*}

Due to the number of FRs in this study, we briefly analyze their feedback by discussing their mean, median, and distribution. We analyze their feedback by involving the related topic together with numerical and open-ended questions. Moreover, combined with their background, we discuss the insights and findings from their reported field experiences.
In \textbf{Q43}, the estimation of SA before an incident was assessed. From the responses, the mean SA was 3.3, and the median was 3.0 (See \cref{fig:SA_semantics_mark}). In \cref{fig: Score Distribution by Role}, it is seen that FRs with a higher position are usually more informed about the incident's SA compared to the operators. Additionally, in \cref{fig: Score Distribution by Experience Level}, a higher score for SA is seen by less experienced operators compared to more years of experience. A potential reason for this difference is that the senior FRs (leaders) need to think more about the overall mission, whereas junior FRs handle specific lower-level tasks that require less SA. I.e. senior FRs might require higher levels of abstract information e.g., high-level semantics and context.


FRs use a variety of methods to obtain SA (\textbf{Q44}). Communication with other people from the site, and in-person exploration, are the most popular ways to obtain information, suggesting that human perception still dominates in obtaining SA. FRs have not yet deeply depended on robot sensors such as cameras. However, such sensors are still welcomed in many deployments and gained 15 votes out of 21. Moreover, a comparison of the current scene against historical data is mentioned. Since this type of information requires support from other departments or individuals and is not always available, it works as a supplement to first-hand information and observations.

The challenges to obtain SA (\textbf{Q45}) come from different aspects of field deployments, including hardware and device malfunction \cite{ramachandran2021resilient}, complex environments, lacking adequate perception, disorganized data, and communication problems with team members.

Based on the feedback of (\textbf{Q46}), we summarize the important semantics in the mission in \cref{fig:important semantics}. If we fit these semantics into the proposed semantic taxonomy in \cite{ruan2022taxonomy}, we find many of them are low-level semantics that can be obtained by robots e.g. location, distance in the blue box. Some of them are high-level semantics e.g. risk in the red box. And many of them are context e.g. medical center, emergency infrastructure in the green box. According to the answers, Junior FRs, who work on the front line and handle the specific operations, focus on the semantics in the blue and red boxes. Senior FRs, who are often the team leader or commander, need to do overall management of the mission. They focus on the semantics and context in the red and green boxes more often.

Results from \textbf{Q47} and \textbf{Q48} show that most FR experienced a slight loss of concentration, fatigue, and cognitive overload during the mission (mean 2.8 on a score from one to five, median 3, see \cref{fig:SA_semantics_mark}). The sources of fatigue and cognitive overload are tasks, environments, and ethical professional constraints. This problem will not be solved by the application of a specific technology, but can be partially alleviated by the use of semantics \cite{ruan2025exploratory}.

Combining the \textbf{Q49}, \textbf{Q50}, \textbf{Q51}, \textbf{Q52} and \textbf{Q53},
about FRs' views on semantic information. Most FRs feel semantics has a positive effect on obtaining SA (mean 3.6 on a score from one to five, median 4, see \cref{fig:SA_semantics_mark}). FRs with a higher education background and experience believe semantics help SA more (See \cref{fig: Score Distribution by Education Level}). Similarly, FRs with higher education level and experience use semantics more in the mission (mean 3.5 on a score from one to five, median 4, see \cref{fig:SA_semantics_mark}). Most FRs have the consensus that semantics help to predict the situation (mean 3.9 on a score from one to five, median 4, see \cref{fig:SA_semantics_mark}). In terms of the delivery methods of semantics, current methods are mostly a combination of vision and sound (from fixed or robot cameras at the base camp). Haptics and vibrations are used in a few special tasks from the joystick, e.g. the deployment of drones. Hence, these answers denote that multimodal is the best option to build SA. FRs who operate the drones are expecting to see an increase in haptics and vibration in other robotic tasks, along with the increasing application of semantics in the future.

\textbf{Q54} indicates FRs have the consensus that AI understanding is valuable in the mission (mean 4 on a score from 1-5, median 4). They probably expect the AI-generated SA to help them reduce the cognitive workload and avoid potential risks.

Based on the feedback from \textbf{Q55} and \textbf{Q56}, the mean accuracy threshold for achieving trust in semantics is 74.6\% (i.e. participants will trust the semantics if it has over 75\% accuracy), and the mean accuracy threshold for usefulness is 67.8\% (i.e. participants will use the semantics if it has over 68\% accuracy). In \cref{fig:accuracy change}, it can be seen that the trust threshold is generally higher than the usefulness threshold for most participants. It indicates that having semantics can be useful even if participants do not feel they can completely trust them. In other words, FRs are eager and expecting to see increasing amounts of semantic information becoming used in a useful way in the field, but it will take longer before these are fully trusted. We further compare the background of FRs who have a higher accuracy threshold for usefulness than for trust. Moreover, some FRs believe the threshold of trust and usefulness in this context is the same concept. I.e., if they are willing to use a certain type of semantic information, it means that they already trust it. We assume the difference might come from variation in the different types of tasks that each participant performs in their work. Some tasks and decisions might have much stricter requirements on data accuracy than others. Hence, these FRs might emphasize the accuracy of the semantics more highly.

We now specifically discuss the connections between the FR's views and their backgrounds, roles in the mission, experience in the field, and education levels.

From \cref{fig: Score Distribution by Role}, we find that different roles in the mission do not cause much variation in opinions. Hence, the role might not be the main reason that differentiates the FR's views. However, different levels of experience in the field do correlate with differences in opinions. In \cref{fig: Score Distribution by Experience Level}, FRs with more than 10 years of field experience have an overall higher rating on Q49, Q50, Q53, and Q54. These questions are related to new technologies for semantic-based SA in future deployments. This suggests that senior FRs are more positive and have higher expectations on this compared with other groups. FRs with four to ten years of experience have a more even distribution on these questions. Junior FRs (less than three years of experience) show much greater uncertainty about their thinking. In the context of education from \cref{fig: Score Distribution by Education Level}, it's interesting that the score on Q43 shows a decline when education levels increase. This suggests that FRs with higher education tend to be cautious before the mission. Moreover, there is a decline in distributions on Q54, suggesting FR's with higher education levels do not show the same level of confidence in AI applications for building SA. They show more independence in AI assistance than other groups. 


\begin{figure}[t]  
    \centering
    \includegraphics[width=0.95\columnwidth]{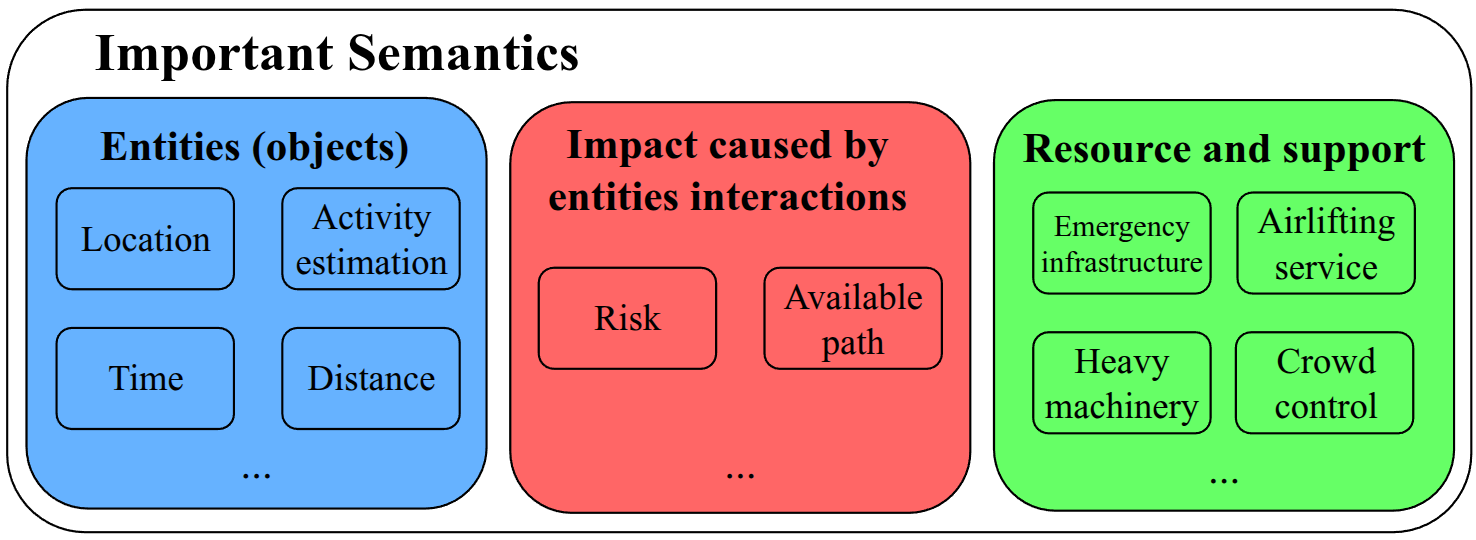}
    \caption{Important semantics for FRs in the mission according to the questionnaire.}
    \label{fig:important_semantics}
\end{figure}

We gained some additional insights from FR comments in the open-ended questions. For instance, one FR highlights the importance of distance to the object (e.g. distance to the gas tank), which could help reduce potential risks. One senior FR with experience in robot deployments, especially drone deployment, feels that the acceptance accuracy for trustworthy and useful semantics varies according to different missions and scenarios. By analysing his background, we found that he was the leader of the squad and participated in many different types of response missions. 

FRs' background of role, experience level, and education level inevitably have correlations. However, it's outside the scope of this paper to explore FRs' views about SA and semantic information. Thus, we did not discuss this topic in detail.

In summary, most FRs across a variety of backgrounds are willing to apply semantics to help them build SA. On one hand, senior FRs have more experience in using semantics to obtain SA, and show higher expectations for using semantics-based SA in the future. On the other hand, they do not feel so dependent on assisted SA compared with other groups. They have enough expert knowledge and experience to leverage information from different sources and mitigate or overcome the impact of errors in the system or incomplete information.

\section{Conclusion}
This study has explored views about SA and semantics in the context of robot deployments of FRs. We found: 

\begin{itemize}
\item \textbf{a)} The most surveyed FRs have positive attitudes towards robots and expectations of more robot deployments. 
\item \textbf{b)} The important semantics in the mission, and some of them could be obtained with the assistance of the AI and robotic techniques, e.g. using images or sensor fusion techniques to obtain semantics such as object identification and positioning, and more abstract situational assessments such as risk. 
\item \textbf{c)} Most FRs are expecting and willing to apply AI-based semantics in their missions, even though it does not achieve a high accuracy and robustness of SA. I.e. they could deploy a robot that does not meet their detection accuracy expectation.
\end{itemize}
Furthermore, such semantic information could be delivered by a well-organized multi-modal way e.g. visual display, and sound notifications in future work.

\section{Acknowledgment}
We greatly appreciate all the FRs who joined our study around the world. They shared their expertise and valuable time with us.

\bibliographystyle{IEEEtran}
\bibliography{references}

@article{senaratne2025framework,
  title={A Framework for Dynamic Situational Awareness in Human Robot Teams: An Interview Study},
  author={Senaratne, Hashini and Tian, Leimin and Sikka, Pavan and Williams, Jason and Howard, David and Kuli{\'c}, Dana and Paris, C{\'e}cile},
  journal={ACM Transactions on Human-Robot Interaction},
  year={2025},
  publisher={ACM New York, NY}
}

@article{joshi2015likert,
  title={Likert scale: Explored and explained},
  author={Joshi, Ankur and Kale, Saket and Chandel, Satish and Pal, D Kumar},
  journal={British journal of applied science \& technology},
  volume={7},
  number={4},
  pages={396},
  year={2015},
  publisher={Sciencedomain International}
}

@article{ruan2025exploratory,
  title={An Exploratory Study on Human-Robot Interaction using Semantics-based Situational Awareness},
  author={Ruan, Tianshu and Ramesh, Aniketh and Stolkin, Rustam and Chiou, Manolis},
  journal={arXiv preprint arXiv:2507.17376},
  year={2025}
}

@Article{Carradore2024,
author={Carradore, Marco
and Artioli, Giovanna
and Sarli, Annavittoria},
title={The General Attitudes Towards Robots Scale (GAToRS): A Preliminary Validation of the Italian Version},
journal={International Journal of Social Robotics},
year={2024},
month={Oct},
day={01},
volume={16},
number={9},
pages={2001-2018},
issn={1875-4805},
doi={10.1007/s12369-024-01170-w},
url={https://doi.org/10.1007/s12369-024-01170-w}
}

@INPROCEEDINGS{yang2014,
  author={Yang, Dongyi and Schafer, James and Wang, Sili and Ganz, Aura},
  booktitle={2014 36th Annual International Conference of the IEEE Engineering in Medicine and Biology Society}, 
  title={Autonomous mobile platform for enhanced situational awareness in Mass Casualty Incidents}, 
  year={2014},
  volume={},
  number={},
  pages={898-901},
  doi={10.1109/EMBC.2014.6943736}}

@INPROCEEDINGS{allison2024,
  author={Allison, Mark and Farmer, Michael and Song, Zheng},
  booktitle={2024 IEEE 14th Annual Computing and Communication Workshop and Conference (CCWC)}, 
  title={Towards Distributed Learning to Support Situational Awareness for Robotic Team Augmented Humanitarian Disaster Response}, 
  year={2024},
  volume={},
  number={},
  pages={0370-0374},
  doi={10.1109/CCWC60891.2024.10427713}}

@ARTICLE{Cooper2013-bq,
  title    = "Measuring situation awareness in emergency settings: a systematic
              review of tools and outcomes",
  author   = "Cooper, Simon and Porter, Joanne and Peach, Linda",
  journal  = "Open Access Emerg Med",
  volume   =  6,
  pages    = "1--7",
  month    =  dec,
  year     =  2013,
  address  = "New Zealand",
  language = "en"
}

@article{endlsey1995,
author = {Mica R. Endsley},
title ={Toward a Theory of Situation Awareness in Dynamic Systems},

journal = {Human Factors},
volume = {37},
number = {1},
pages = {32-64},
year = {1995},
doi = {10.1518/001872095779049543},

URL = { 
    
        https://doi.org/10.1518/001872095779049543
    
    

},
eprint = { 
    
        https://doi.org/10.1518/001872095779049543
    
    

}
}

@inproceedings{betta2024perceptions,
  title={Perceptions and opinions of rescuers about a quadruped robot in an earthquake scenario},
  author={Betta, Zoe and Gaudino, Alessandro and Benini, Alessandro and Recchiuto, Carmine Tommaso and Sgorbissa, Antonio},
  booktitle={2024 33rd IEEE International Conference on Robot and Human Interactive Communication (ROMAN)},
  pages={1092--1099},
  year={2024},
  organization={IEEE}
}

@article{murphy2025rural,
  title={Rural hospital incident command leaders’ perceptions of disaster preparedness},
  author={Murphy, Jason P and Bergstr{\"o}m, Clara and Gyllencruetz, Lina},
  journal={BMC Emergency Medicine},
  volume={25},
  number={1},
  pages={45},
  year={2025},
  publisher={Springer}
}

@article{koverola2022general,
  title={General attitudes towards robots scale (GAToRS): A new instrument for social surveys},
  author={Koverola, Mika and Kunnari, Anton and Sundvall, Jukka and Laakasuo, Michael},
  journal={International Journal of Social Robotics},
  volume={14},
  number={7},
  pages={1559--1581},
  year={2022},
  publisher={Springer}
}

@inproceedings{snyder2019situational,
  title={Situational awareness enhanced through social media analytics: A survey of first responders},
  author={Snyder, Luke S and Karimzadeh, Morteza and Stober, Christina and Ebert, David S},
  booktitle={2019 IEEE International Symposium on Technologies for Homeland Security (HST)},
  pages={1--8},
  year={2019},
  organization={IEEE}
}

@INPROCEEDINGS{ruan2022taxonomy,
  author={Ruan, Tianshu and Wang, Hao and Stolkin, Rustam and Chiou, Manolis},
  booktitle={2022 IEEE International Symposium on Safety, Security, and Rescue Robotics (SSRR)}, 
  title={A Taxonomy of Semantic Information in Robot-Assisted Disaster Response}, 
  year={2022},
  volume={},
  number={},
  pages={285-292},
  doi={10.1109/SSRR56537.2022.10018727}}

@inproceedings{huang2023went,
  title={What went wrong? closing the sim-to-real gap via differentiable causal discovery},
  author={Huang, Peide and Zhang, Xilun and Cao, Ziang and Liu, Shiqi and Xu, Mengdi and Ding, Wenhao and Francis, Jonathan and Chen, Bingqing and Zhao, Ding},
  booktitle={Conference on Robot Learning},
  pages={734--760},
  year={2023},
  organization={PMLR}
}

@article{ramachandran2021resilient,
  title={Resilient monitoring in heterogeneous multi-robot systems through network reconfiguration},
  author={Ramachandran, Ragesh Kumar and Pierpaoli, Pietro and Egerstedt, Magnus and Sukhatme, Gaurav S},
  journal={IEEE Transactions on Robotics},
  volume={38},
  number={1},
  pages={126--138},
  year={2021},
  publisher={IEEE}
}

@INPROCEEDINGS{10558013,
  author={Tzoumas, Georgios and Salinas, Lucio and McConville, Alex and Richardson, Tom and Hauert, Sabine},
  booktitle={2024 IEEE International Conference on Advanced Robotics and Its Social Impacts (ARSO)}, 
  title={Use case design for swarms of firefighting UAVs via mutual shaping}, 
  year={2024},
  volume={},
  number={},
  pages={43-48},
  doi={10.1109/ARSO60199.2024.10558013}}

@inproceedings{mcconville2024adoption,
  title={Adoption of uav swarm technology: Survey and opinions of firefighters},
  author={McConville, Alex and Tzoumas, Georgios and Salinas, Lucio R and Munera, Marcela and Hauert, Sabine},
  booktitle={2024 IEEE International Conference on Advanced Robotics and Its Social Impacts (ARSO)},
  pages={228--234},
  year={2024},
  organization={IEEE}
}

\end{document}